\documentclass[sigconf]{acmart}

\AtBeginDocument{%
  \providecommand\BibTeX{{%
    \normalfont B\kern-0.5em{\scshape i\kern-0.25em b}\kern-0.8em\TeX}}}

\usepackage{multirow}
\usepackage{multicol}
\usepackage{booktabs} % To thicken table lines
\usepackage{graphicx}
\usepackage{amsmath}

\usepackage{amssymb}% http://ctan.org/pkg/amssymb
\usepackage{pifont}% http://ctan.org/pkg/pifont

\DeclareMathOperator*{\argmax}{arg\,max}

\newcommand{\cmark}{\ding{51}}%
\newcommand{\xmark}{\ding{55}}%
\newcommand{\ie}{\textit{i}.\textit{e}. }
\newcommand{\eg}{\textit{e}.\textit{g}. }
\newcommand{\etal}{\textit{et al.}}

\usepackage{subfigure}
\usepackage[capitalize]{cleveref}
\crefname{section}{Sec.}{Secs.}
\Crefname{section}{Section}{Sections}
\Crefname{table}{Table}{Tables}
\crefname{table}{Tab.}{Tabs.}

\usepackage{hyperref}

\hypersetup{
    colorlinks=true,
    linkcolor=blue,
    citecolor=blue,      
    urlcolor=blue,
}

\copyrightyear{2022}
\acmYear{2022}
\setcopyright{acmcopyright}
\acmConference[SIGIR '22]{Proceedings of the 45th International ACM SIGIR Conference on Research and Development in Information Retrieval}{July 11--15, 2022}{Madrid, Spain}
\acmBooktitle{Proceedings of the 45th International ACM SIGIR Conference on Research and Development in Information Retrieval (SIGIR '22), July 11--15, 2022, Madrid, Spain}
\acmPrice{15.00}
\acmDOI{10.1145/3477495.3532083}
\acmISBN{978-1-4503-8732-3/22/07}

\begin{document}
\fancyhead{} 

\title{You Need to Read Again: Multi-granularity Perception Network for Moment Retrieval in Videos}

\author{Xin Sun}
\authornote{Work done during an internship at CloudWalk Technology.}
\affiliation{%
  \institution{Shanghai Jiao Tong University}
     \city{Shanghai}
  \country{China}
}
\email{huntersx@sjtu.edu.cn}

\author{Xuan Wang}
\affiliation{%
  \institution{Peking University}
     \city{Beijing}
  \country{China}
}
\email{xuan.wang@pku.edu.cn}

\author{Jialin Gao}
\authornotemark[1]
\affiliation{%
  \institution{Shanghai Jiao Tong University}
  \city{Shanghai}
  \country{China}
}
\email{ jialin_gao@sjtu.edu.cn}

\author{Qiong Liu}
\affiliation{%
 \institution{CloudWalk Technology Co., Ltd}
    \city{Shanghai}
  \country{China}
 }
\email{liuqiong@cloudwalk.com}

\author{Xi Zhou}
\affiliation{%
  \institution{CloudWalk Technology Co., Ltd}
     \city{Shanghai}
  \country{China}
 }
\email{zhouxi@cloudwalk.com}

\begin{abstract}
Moment retrieval in videos is a challenging task that aims to retrieve the most relevant video moment in an untrimmed video given a sentence description. Previous methods tend to perform self-modal learning and cross-modal interaction in a coarse manner, which neglect fine-grained clues contained in video content, query context, and their alignment. 
To this end, we propose a novel \textbf{M}ulti-\textbf{G}ranularity \textbf{P}erception \textbf{N}etwork (\textbf{MGPN}) that perceives intra-modality and inter-modality information at a multi-granularity level. Specifically, we formulate moment retrieval as a multi-choice reading comprehension task and integrate human reading strategies into our framework. A coarse-grained feature encoder and a co-attention mechanism are utilized to obtain a preliminary perception of intra-modality and inter-modality information. 
Then a fine-grained feature encoder and a conditioned interaction module are introduced to enhance the initial perception inspired by how humans address reading comprehension problems.
Moreover, to alleviate the huge computation burden of some existing methods, we further design an efficient choice comparison module and reduce the hidden size with imperceptible quality loss. Extensive experiments on Charades-STA, TACoS, and ActivityNet Captions datasets demonstrate that our solution outperforms existing state-of-the-art methods. 
% Our source code is available at \href{https://github.com/Huntersxsx/MGPN}{\textcolor{blue}{Huntersxsx/MGPN}}.
Codes are available at \href{https://github.com/Huntersxsx/MGPN}{\textcolor{blue}{github.com/Huntersxsx/MGPN}}.
\end{abstract}




\maketitle
\begin{figure}[t]
    \centering
    \includegraphics[width=1.0\linewidth]{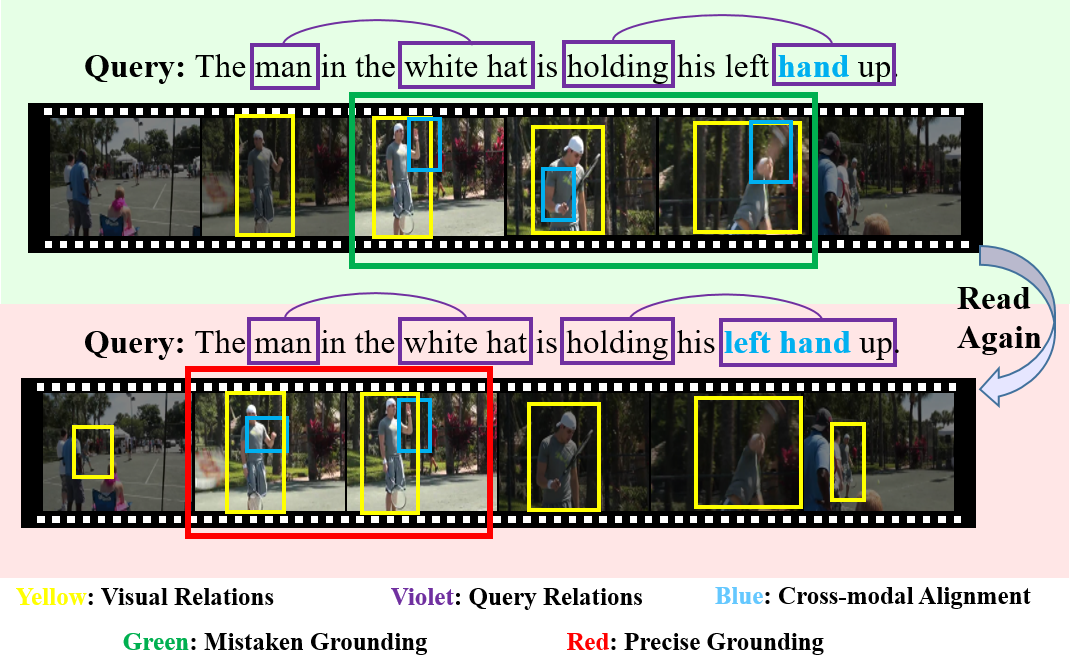}
    \caption{Illustration of moment retrieval and our motivation. \textbf{Upper:} Existing works tend to encode intra-modality and inter-modality information in a coarse-grained manner (with one-step reasoning), which may neglect semantic details (\eg "left" hand) and lead to misalignment. \textbf{Bottom}: Inspired by human's reading strategies, our MGPN perceives intra-modality  and inter-modality information at a multi-granularity level (with deeper reasoning). 
    % It's best to zoom in and view in colour.
    }
    \label{fig:motivation}
\end{figure}
\section{Introduction}
Tremendous videos over the internet contain diverse human activities, which are beneficial for us to perceive the real world. This phenomenon has motivated many popular tasks such as action recognition \cite{su2020convolutional, gao2020accurate, gao2021skeleton}, video retrieval \cite{yu2018joint, gabeur2020multi}, to name a few. However, most online videos are long, untrimmed and contain irrelevant content, thus automatic video content analysis methods are in emergent need. Temporal Action Localization \cite{shou2016temporal,lin2018bsn} is a task to detect action instances in untrimmed videos while it neglects the fact that videos are often accompanied by sentence descriptions. Gao \etal \cite{gao2017tall}, and Hendricks \etal \cite{anne2017localizing} thus take full advantage of annotated language information and propose the Moment Retrieval task. Given an untrimmed video and a sentence description, the goal of this task is to retrieve the video moment most semantically related to the description. It is a meaningful yet challenging task because it is required to simultaneously understand complicated video content and sentence context.

Existing works for this task generally fall into two categories: two-stage models \cite{gao2017tall,anne2017localizing,xu2019multilevel} and one-stage models \cite{lu2019debug, Zeng_2020_CVPR, zhang2020span}. Two-stage methods usually follow a propose-then-rank pipeline and achieve more decent performance than one-stage methods. As shown in Figure \ref{fig:motivation}, most two-stage models \cite{gao2017tall,xiao2021boundary, yuan2019semantic} learn self-modal relations and cross-modal interaction in a coarse-grained manner. They tend to use two separate feature encoders to capture intra-modality information, and then utilize attention mechanism to explore inter-modality information. 
However, these two-stage methods suffer from two notorious drawbacks: (i) They neglect detailed information in intra-modality and inter-modality (\eg some subtle objects boxed by yellow anchors in Figure \ref{fig:motivation} bottom part while ignored in the upper part; key information like "left" that are crucial for accurate alignment), which needs to be sufficiently perceived with a deeper reasoning stage. (ii) In order to achieve a high recall, two-stage models are required to densely generate candidate moments, which is computation-consuming and time-consuming.

Considering the promising performance obtained by existing two-stage methods, we also follow the propose-then-rank pipeline to tackle moment retrieval task. For the drawbacks mentioned above, we apply a coarse-to-fine strategy to mine intra-modality and inter-modality information at a multi-granularity level. Both moment content and boundary information are taken into account for comprehensive interaction. As for the large model size of previous two-stage methods, we further design a lightweight choice comparison module and reduce the hidden size of our model without apparent performance decrease. Most similar to our work is SMIN \cite{wang2021structured}, which also explores cross-modal information in a coarse-to-fine manner. However, the interaction module of SMIN is elaborated and sophisticated, in which cross-modal interaction and moment interaction are simultaneously exploited in an iterative way.

In this paper, we formulate Moment Retrieval task as a Multi-choice Reading Comprehension problem, and then we learn from some strategies used in reading comprehension. Some published works \cite{guthrie1987literacy,zheng2019human} have researched human behavior in reading comprehension. Recent work \cite{zhang2021retrospective} has demonstrated that the reread strategy commonly used by human is beneficial for machine reading comprehension task. They claimed that reading comprehension is a two-stage process, where we need to first read through the passage and question to grasp the general idea, and then reread the full text to verify the answer. In our study, we follow several human reading habits \cite{guthrie1987literacy,zheng2019human,zhang2021retrospective}, where people first read text passage and questions roughly to obtain a preliminary perception of each answer. Then they thoroughly read passage and questions again for deeper reasoning. Finally people compare all the answers carefully to make correct decisions. Inspired by such a pipeline utilized in multi-choice problems, we integrate several reading strategies into our model: (i) passage question reread that finds fine-grained intra-modality clues to thoroughly understand video content and query context; (ii) enhanced passage question alignment that fuses video and query once again to explore deeper inter-modality information; (iii) choice comparison that encodes the interaction information among candidate moments to empower our model the ability of distinguishing similar video moments. Our main contributions can be summarized as follows:
\begin{itemize}
    \item We formulate Moment Retrieval task as a Multi-choice Reading Comprehension problem and integrate several human reading strategies into our proposed Multi-granularity Perception Network (MGPN).
    \item We propose a novel fine-grained feature encoder and a conditioned interaction module to fully mine semantic information in both intra-modality and inter-modality. 
    \item We design an efficient choice comparison module and reduce the hidden size of our model with imperceptible quality loss to alleviate the high computational complexity of existing two-stage methods.
    \item Experiments conducted on three popular benchmarks demonstrate that our MGPN outperforms the existing state-of-the-art methods.
\end{itemize}
\section{Related Work}

\begin{figure*}[t]
    \centering
    \includegraphics[width=1.0\linewidth]{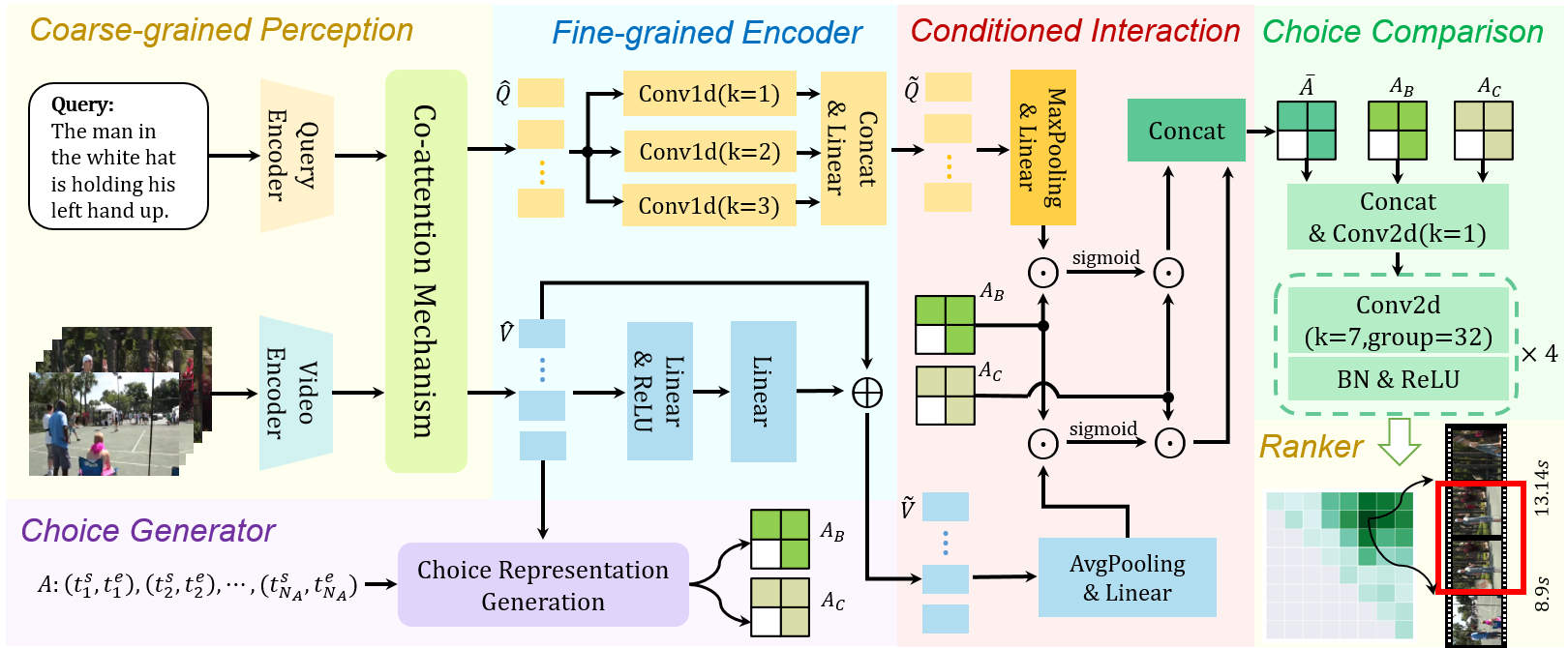}
    \caption{The framework of our proposed Multi-granularity Perception Network. The overall procedure is in line with human reading habits. First we utilize a coarse-grained feature encoder and a co-attention module to obtain a preliminary perception of intra-modality and inter-modality information. Then we generate both content-level feature map $\mathbf{A_C}$ and boundary-level feature map $\mathbf{A_B}$. Next, a fine-grained feature encoder and a conditioned interaction module are applied to enhance the understanding. Finally we capture adjacent temporal relations among moments with a choice comparison module and retrieve the most relevant moment with a choice ranker.
    }
    \label{fig:fw}
\end{figure*}

\noindent
\textbf{Moment Retrieval in videos.} This task aims at predicting the start and end time of the described activity given an untrimmed video and a language query, which was introduced by \cite{anne2017localizing,gao2017tall}. Existing methods on Moment Retrieval can be grouped into two categories, \ie two-stage methods and one-stage methods.

Most two-stage methods \cite{gao2017tall,anne2017localizing,xu2019multilevel,xiao2021boundary,gao2021relation} tend to follow a propose-then-rank pipeline, where they first generate a number of candidate moments and then rank them relying on their matching scores, the highest score moment is selected as the retrieval result. Gao \etal \cite{gao2017tall} and Hendricks \etal \cite{anne2017localizing} predefined candidate moments with sliding windows and calculate the similarity with language queries in a joint embedding space. Because the performance of two-stage methods highly relies on the predefined moments, some works seek to improve their quality. Liu \etal \cite{liu2018attentive} introduced a temporal memory attention network to memorize the contextual information for each moment. Chen \etal \cite{chen2018temporally} established frame-by-word interactions to obtain token-aware video representation. To further distinguish visually similar moments, MAN \cite{zhang2019man} and 2D-TAN \cite{zhang2020learning} modeled complex moment-wise temporal relations with iterative graph or stacked convolution layers. Recently, Gao \etal \cite{gao2021relation} utilized a sparsely-connected graph attention module to collect contextual information from adjacent moments. However, two-stage methods need to sample video moment candidates densely for more precise predicted timestamps, which leads to heavy computation cost and lack of flexibility.

One-stage methods \cite{lu2019debug,Zeng_2020_CVPR,zhang2020span,li2021proposal,rodriguez2020proposal} avoid the need of predefined candidate moments, which can obtain moments with flexible length and perform more efficiently. They directly predict the probability for each frame whether it is the boundary frame. Yuan \etal \cite{yuan2019find} regressed the temporal coordinates with a multi-modal co-attention mechanism. Lu \etal \cite{lu2019debug} regarded all frames falling in the ground truth segment as foreground and made full use of 
positive samples to alleviate the severe imbalance problem. Zeng \etal \cite{Zeng_2020_CVPR} leveraged dense supervision from the sparse annotations to regress the distances from each frame to the boundary frame. However, one-stage methods usually have poor performance because they neglect rich moment-level information which is important for precise localization.

\noindent
\textbf{Machine Reading Comprehension.} This task aims at predicting appropriate answers given a text passage and a relevant question, which is a challenging task in natural language understanding domain \cite{hermann2015teaching, wang2019evidence}. In view of the form of answers, Machine Reading Comprehension (MRC) has two common variants \ie span extraction and multi-choice selection. The former needs to locate a certain span that seems most related to the given question \cite{rajpurkar2016squad, rajpurkar2018know} while the latter is required to select a correct answer in a given candidate choices set\cite{lai2017race,sun2019dream}. Moment Retrieval in videos task can be viewed as  Machine Reading Comprehension task when we treat the video as a text passage, language query as a question description and video candidate moments as a list of answer choices, respectively. Some published works\cite{lu2019debug, ghosh2019excl, zhang2020span} regarded the moment retrieval task as a span-based MRC and directly regressed the start and end time. While Gao \etal \cite{gao2021relation} regarded it as a multi-choice MRC and used propose-then-rank pipeline to tackle it. 

Although some existing works are aware of the relevance between moment retrieval task and machine reading comprehension task, they simply pay attention to the task format while neglect common strategies used in reading comprehension. Reading passage once again is a common reading behavior and proved to be an effective strategy for MRC task \cite{guthrie1987literacy,zheng2019human}. Zhang \etal \cite{zhang2021retrospective} was motivated by human reading habits and proposed a retrospective reader to tackle complex reading comprehension with reading and verification strategies. Inspired by these promising works, we propose a Multi-granularity Perception Network that integrates human reading strategies including passage question reread, enhanced passage question alignment and choice comparison to empower our model the ability of thoroughly understanding video and query contents.

\section{The Proposed Method}

\subsection{Overview}
Given an untrimmed video $\mathbf{V}$ and a language query $\mathbf{Q}$ which describes a moment in $\mathbf{V}$, Moment retrieval task aims to retrieve the most relevant temporal moment with a start and end time point $(t^s,t^e)$ from a candidate moments set $\mathbf{A}$. To some extent, it is similar with the Multi-choice Reading Comprehension task in which the input triplet $(\mathbf{V,Q,A})$ denotes text passage, question description and candidate answers, respectively. With these notations, we can formulate the purpose of moment retrieval task as:
\begin{equation}
    \begin{aligned}
        \argmax_{n}P(a_{n}|(\mathbf{V}, \mathbf{Q}, \mathbf{A})).\\
        % s.t \quad 0<i<j<v \quad
    \end{aligned}
    \label{eq:se}
\end{equation}

Other than only retrieving the most possible moment, many previous works in moment retrieval \cite{liu2018attentive,zhang2019cross} also report their retrieval score in terms of top-$K$ most matching candidates. We follow them and predict $\{(p_n, t^s_n, t^e_n)\}_{n=1}^K$ for fair comparison, where $p_n, t^s_n, t^e_n$ represent the retrieval probability, start and end time of the answer $a_n$, respectively. Formally, we denote the input video as $\mathbf{V}=\{v_t\}^{T_V}_{i=t}$, and the language query as $\mathbf{Q}=\{q_n\}^{N_Q}_{n=1}$, where $v_t$ is the $t$-th frame in the video and $q_n$ is the $n$-th word in the sentence. $T_V$ and $N_Q$ represent the length of the video and sentence, respectively. Following the sparse sampling strategy proposed in Zhang \etal \cite{zhang2020learning}, we construct a series of candidate moments as the answer set $\mathbf{A}=\{a_n\}^{N_A}_{n=1}$, where $N_A$ is the total number of valid candidates and $a_n$ represents one possible candidate moment. Without additional mention, the video moment and answer/choice are interchangeable in our paper.

In this section, we will introduce our framework MGPN as shown in Figure~\ref{fig:fw}. Our model totally consists of seven components: (i) coarse-grained feature encoder; (ii) multi-modal co-attention module; (iii) candidate moments generation module; (iv) fine-grained feature encoder; (v) conditioned interaction module; (vi) choice comparison module; (vii) choice ranker. Specifically, we first feed the video and sentence embeddings into a coarse-grained encoder and compound them to obtain a preliminary aligned representation. There followed a candidate representation generation module to build both content-level and boundary-level moment representations. And then, we follow the reread strategy used in reading comprehension and encode video and query features deeply to get fine-grained intra-modality representations. Next step we combine them with generated moment features to obtain enhanced aligned moment features. We further apply a choice comparison module to help distinguish similar moments. Finally, we retrieve the most relevant video moment in the candidate set through a choice ranker.

\subsection{Coarse-grained Feature Encoder}
\label{sec:method-coarse-encoder}
\noindent
\textbf{Video Encoding.} Given an input video $\mathbf{V}=\{v_t\}^{T_V}_{i=t}$, we first extract the clip-wise features by a pre-trained network (\eg C3D \cite{tran2015learning}, VGG \cite{simonyan2014very} or I3D \cite{carreira2017quo}), then we apply a temporal 1D convolution layer and an average pooling layer to map the clip-wise features into a desired dimension $C$ and length $T$. Considering the sequential characteristic in videos, we further employ a bi-directional GRU \cite{chung2014empirical} to capture the temporal dependencies among video clips. The output of video encoder can be represented as $\bar{\mathbf{V}}\in \mathbb{R} ^{T\times C}$.\\
\noindent
\textbf{Query Encoding.} Given a language query $\mathbf{Q}=\{q_n\}^{N_Q}_{n=1}$, we first encode each word with a pre-trained 300-dimensional Glove model\cite{pennington2014glove}. Then we sequentially feed the initialized embeddings into a bi-directional GRU to capture the contextual information in sentences. The output feature representation of the input sentence can be denoted as $\bar{\mathbf{Q}}\in \mathbb{R} ^{L\times C}$, where $L$ is the length of the longest sentence in a mini-batch.

\subsection{Multi-modal Co-attention Module}
\label{sec:method-coarse-fuse}
% The modality-wise encoder aims to capture the rich intra-modality context between each snippet or token.
This module aims to preliminarily align the encoded features of two modalities to obtain query-aware video features $\hat{\mathbf{V}}$ and video-aware query features $\hat{\mathbf{Q}}$. We apply a multi-modal co-attention mechanism to capture the inter-modality context. Specially, we first feed the encoded query features $\bar{\mathbf{Q}}$ into a linear layer and apply a $\mathit{softmax}$ function to get query attention weights $\mathbf{a}^Q$. Each single element $a_j^Q$ in $\mathbf{a}^Q$ represents the importance of the $\mathit{j}$th token. We then sum up each token's feature based on $a_j^Q$ and obtain the attended query features $\mathbf{q}^{attn}$. At last, query-aware video features are achieved by attending to the video based on $\mathbf{q}^{attn}$ and utilizing a $\ell _2$ normalization. The fusion procedure can be written as:
\begin{equation}
\begin{gathered}
% \begin{split}
         \mathbf{a}^Q = \mathit{softmax}(\mathbf{W}_Q \bar{\mathbf{Q}}^T + \mathbf{b}_Q) \in \mathbb{R} ^{1\times L}\\
         \mathbf{q}^{attn} = \sum a_j^Q \bar{\mathbf{q}}_j \in \mathbb{R} ^{1\times C}\\
         \hat{\mathbf{V}} = \left \| \mathbf{q}^{attn} \odot \bar{\mathbf{V}} \right \|_F \in \mathbb{R} ^{T\times C}
% \end{split}
\end{gathered}
\end{equation} 
where $\mathbf{W}_Q$ and $\mathbf{b}_Q$ are learnable parameters, $\bar{\mathbf{q}}_j$ is the feature of $\mathit{j}$th token, $\odot$ and $\Vert \cdot \Vert_F$ denote Hadamard product and $\ell _2$ normalization, respectively. Due to the symmetry of multi-modal interaction, we can achieve video-aware query features $\hat{\mathbf{Q}} \in \mathbb{R} ^{L\times C}$ in a similar way. 

\subsection{Choice Representation Generation}
\label{sec:method-generator}
\begin{figure}[t]
    \centering
    \includegraphics[width=1.0\linewidth]{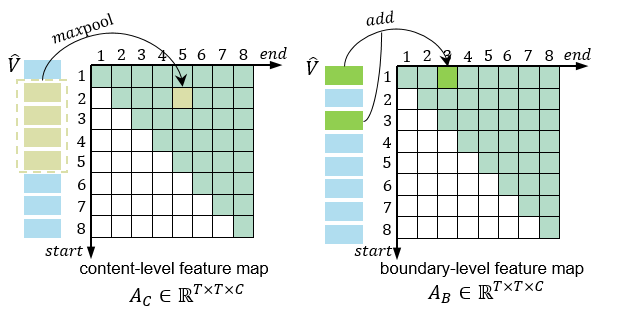}
    \caption{
    \textbf{Left:} We use MaxPooling operation to generate content-level feature map. \textbf{Right:} We use Addition operation to generate boundary-level feature map. 
    }
    \label{fig:gen}
\end{figure}

Following previous works \cite{zhang2020learning, gao2021relation}, we construct a two-dimensional temporal map to generate our candidate moments. As shown in Figure \ref{fig:gen}, the vertical axis and horizontal axis respectively represent the start and end clip indices, and each block $(i, j)$ represents a candidate moment from time $i\tau$ to $(j+1)\tau$. We follow the sparse sampling strategy proposed in \cite{zhang2020learning}, thus we can obtain $T \times T$ candidate moments totally while only $N_A$ in them are valid. The lower triangular part of the 2D map is invalid because the start indices of these blocks surpass their end indices, and parts of the upper triangular are masked for the computation consideration. We denote the generated candidate choices set as $\mathbf{A}=\{a_n\}^{N_A}_{n=1}$, where each choice $a_n$ represents a valid moment span from $t^s_n$ to $t^e_n$. 

To roundly capture the temporal correlation in videos, we construct both content-level and boundary-level representations for each moment. Inspired by Zhang \etal \cite{zhang2020learning} and Gao \etal \cite{gao2021relation}, we use \textit{MaxPooling} operation and \textit{Addition} operation to generate content-level moment features $\mathbf{A_C}$ and boundary-level moment features $\mathbf{A_B}$, respectively. Specially, for a candidate moment $a_n$ in set $\mathbf{A}$, we max-pool its corresponding clip features $\mathbf{\hat{v}}_{t^s_n},\dots , \mathbf{\hat{v}}_{t^e_n}$ in $\mathbf{\hat{V}}$ across a time span $(t^s_n, t^e_n)$ to obtain its content-level feature $\mathbf{f^c}_n$; we simply add its start clip feature $\mathbf{\hat{v}}_{t^s_n}$ and end clip feature $\mathbf{\hat{v}}_{t^s_n}$ to get its boundary-level feature $\mathbf{f^b}_n$. The moment features construction strategy can be written as:
\begin{equation}
\begin{gathered}
% \begin{split}
        \mathbf{f^c}_n = MaxPooling(\hat{v}_{t^s_n},\dots , \hat{v}_{t^e_n})\\
        \mathbf{f^b}_n = Addition(\hat{v}_{t^s_n}, \hat{v}_{t^e_n})
% \end{split}
\end{gathered}
\end{equation} 

Although only $N_A$ candidate moments are valid, we maintain invalid candidate moments and zero-pad their features for convenience. Therefore, we can obtain content-level features and boundary-level features of all candidate moments:
\begin{equation}
\begin{gathered}
        \mathbf{A_C}=\{\mathbf{f^c}_n\}^{N_A}_{n=1} \in \mathbb{R} ^{T \times T \times C}\\
        \mathbf{A_B}=\{\mathbf{f^b}_n\}^{N_A}_{n=1} \in \mathbb{R} ^{T \times T \times C}
\end{gathered}
\end{equation}

We suppose that both content-level and boundary-level moment representations are crucial for better retrieval, and we do an investigation in section \ref{sec:generator-compare} to confirm this point.

\subsection{Fine-grained Feature Encoder}
\label{sec:method-fine-encoder}
In the early coarse-grained feature encoders, we use a Bi-GRU operation to briefly encode the content of video and language query. However, this process can only grasp the general information while neglects the rich contextual information hidden in the more detailed level (\eg token-level features), thus we need further encoding to enhance intra-modality representations. When solving multi-choice reading comprehension task, reading passage and questions carefully once again after first reading is a commonly used strategy. Mimicking humans, we add this module in our framework to extract fine-grained intra-modality information. 
% and facilitate better interaction of two modalities. 

\noindent
\textbf{Video Encoding.} Feed-forward network (FFN) is a commonly seen feature encoder which can perceive potential relations among all the feature units. Given the query-aware video features $\hat{\mathbf{V}}$, we design a Residual-FFN to explore fine-grained visual clues for better interaction and maintain the coarse-grained features as the preliminary perception. The procedure can be written as: 
\begin{equation}
\begin{gathered}
        \Tilde{\mathbf{V}} = Linear(ReLU(Linear(\hat{\mathbf{V}}))) + \hat{\mathbf{V}}
\end{gathered}
\end{equation}

The implementation of our video encoder is plain for the sake of retrieval efficiency. Although it can bring some performance improvements as shown in our ablation studies, it may not be convincing enough. our future work will mainly focus on devising a more effective fine-grained video encoder.

\noindent
\textbf{Query Encoding.} As for the sentence sequence, we further explore token-level and phrase-level features for a fine-grained representation. Specially, we first apply a point-wise 1D convolution operation to capture token-level feature, which is denoted as unigram feature $\hat{\mathbf{Q}}^u$. To fully mine the semantic information, we also capture phrase-level feature, which is obtained by applying two convolution operations with different kernel sizes. We group adjacent tokens as a bigram or a trigram, and utilize temporal 1D convolution operations to obtain bigram feature $\hat{\mathbf{Q}}^b$ and trigram feature $\hat{\mathbf{Q}}^t$. 
Finally, we concatenate these three features and feed them into a fully connected layer to integrate them and obtain fine-grained query features $\Tilde{\mathbf{Q}}$: 
\begin{equation}
\begin{gathered}
        \Tilde{\mathbf{Q}} = Linear(Concat(\hat{\mathbf{Q}}^u, \hat{\mathbf{Q}}^b, \hat{\mathbf{Q}}^t))
\end{gathered}
\end{equation}

Through fine-grained feature encoders, we can follow the coarse-to-fine pipeline commonly seen in reading comprehension, and obtain fine-grained video features $\Tilde{\mathbf{V}} \in \mathbb{R} ^{T\times C}$ as well as query features $\Tilde{\mathbf{Q}} \in \mathbb{R} ^{L\times C}$. This human-like strategy is rational and beneficial to the later fusion and retrieval, which is confirmed in our in-depth ablation studies \ref{sec:component} and \ref{sec:plug}. 

\subsection{Conditioned Interaction Module}
\label{sec:method-fine-fuse}
After obtaining fine-grained intra-modality features, we need to compound them with moment-wise features to further enhance the inter-modality feature representation. In this module, we propose a conditioned interaction with gate mechanism to thoroughly extract explicit relations between two modalities.

We set up a symmetry interaction between video and sentence, thus this module is divided into two branches (\ie query-aware branch, video-aware branch). In each branch, we take advantage of a designed gate mechanism to separately learn the query-aware moment representation and video-aware moment representation. Take query-aware branch for example, we first apply \textit{MaxPooling} on the query representation to find the most contributed tokens at each feature dimension and utilize a fully connected layer to map it into moment-level feature space. Boundary-level moment feature emphasizes the boundary information of a moment thus plays an important role in determining the start and end timestamps. Therefore, we combine the boundary-level moment feature $\mathbf{A_B}$ with the transferred query feature $\Tilde{\mathbf{Q}}^{\prime}$ and use \textit{sigmoid} as a gate function to capture semantically correlated boundary information. We further aggregate the gated query feature $\mathbf{G}_Q$ and content-level moment feature $\mathbf{A_C}$ to explore the query-related information contained in candidate moments. The whole interaction procedure of query-aware branch can be formulated as:

\begin{equation}
\begin{gathered}
        \Tilde{\mathbf{Q}}^{\prime} = Linear(MaxPooling(\Tilde{\mathbf{Q}})) \in \mathbb{R} ^{1 \times C}\\
        \mathbf{G}_Q = \sigma(\mathbf{A_B} \odot \Tilde{\mathbf{Q}}^{\prime} ) \in \mathbb{R} ^{1 \times C} \\ 
        \bar{\mathbf{A}}_1 = \mathbf{G}_Q \odot \mathbf{A_C} \in \mathbb{R} ^{T \times T \times C}
\end{gathered}
\end{equation}

Due to the symmetry of our designed interaction module, we then process video-aware branch in a similar way and obtain the video-aware moment features denoted as $\bar{\mathbf{A}}_2$. At last, we integrate these two conditioned features and obtain the fine-grained aligned moment features $\bar{\mathbf{A}}$:

\begin{equation}
\begin{gathered}
        \bar{\mathbf{A}}_2 = \sigma(\mathbf{A_B} \odot Linear(AvgPooling(\Tilde{\mathbf{V}}))) \odot \mathbf{A_C} \\
        \bar{\mathbf{A}} = Concat(\bar{\mathbf{A}}_1, \bar{\mathbf{A}}_2) \in \mathbb{R} ^{T \times T \times 2C}
\end{gathered}
\end{equation}

\subsection{Choice Comparison Module}
\label{sec:method-comparison}
When doing a reading comprehension task, human tends to compare candidate choices carefully before making a selection. Some existing Multi-choice Reading Comprehension methods \cite{ran2019option, zhang2020dcmn+} were inspired by such human behaviour and encoded the comparison information among answer choices to make reasoning more efficient. This reading strategy can also be integrated in moment retrieval task due to its comparability with reading comprehension task. Recent solutions \cite{zhang2019man, zhang2020learning, gao2021relation} in moment retrieval were aware of the importance of adjacent temporal relation among different candidate moments, and utilized stacked convolution layers or graph convolution network to construct it. In our implement. we also apply a choice comparison module to capture discriminative features of different candidate moments, which can help our network distinguish those visually similar video moments for precise localization.

Inspired by 2D-TAN \cite{zhang2020learning}, our choice comparison module also consists of several simple convolution blocks. However, our implementation is more efficient because of two improvements. On one hand, we use group convolution in our blocks instead of vanilla convolution layer used in 2D-TAN \cite{zhang2020learning}. One the other hand, the hidden size of our implementation is set to 256, which is only half of common solutions. Concretely, we first concatenate the fine-grained aligned feature map $\bar{\mathbf{A}}$, content-level feature map $\mathbf{A_C}$ and boundary-level feature map $\mathbf{A_B}$ to integrate their correlations. Then we use a skip-connection and apply a convolution operation with \textit{ReLU} activation function to obtain enhanced fused features $\hat{\mathbf{A}}$. Afterwards, we totally stack four convolution blocks over the fused features to perceive more contextual information from adjacent candidate moments. Each convolution block consists of a group convolution layer followed by a batch normalization and \textit{ReLU} activation function. 
The detailed settings of our convolution blocks are reported in section \ref{sec:details}. The adjacent relations learning process can be briefly formulated as:

\begin{equation}
\begin{gathered}
        \hat{\mathbf{A}} = ReLU(Conv2d(Concat(\bar{\mathbf{A}}, \mathbf{A_B}, \mathbf{A_C}) + \bar{\mathbf{A}}))\\
        \Tilde{\mathbf{A}} = StackedConv2d(\hat{\mathbf{A}}) \in \mathbb{R} ^{T\times T \times C}
\end{gathered}
\end{equation}

After fusing moment features with comparison information across all candidates, we can capture complex temporal dependencies among different candidate moments. Moreover, the effectiveness and efficiency of different choice comparison modules are further discussed in ablation study \ref{sec:experiment-choice-comparison}.

\subsection{Choice Ranker}
\label{sec:method-ranker}
In this module we will generate a retrieval score for each candidate moment and rank them to make decisions. After capturing the relations among different candidate moments, we feed the relation-aware feature $\Tilde{\mathbf{A}}$ into a convolution layer and employ a sigmoid function to generate matching scores for all candidate moments, which can be written as :
\begin{equation}
    \mathbf{P_A} = \sigma(Conv( \Tilde{\mathbf{A}} )),
\end{equation}
where $\sigma$ represents the sigmoid activation function and $\mathbf{P_A} = \{p_n\}^{N_A}_{n=1}$, in which each $p_n$ denotes the probability of the candidate moment $a_n$ being retrieved as the best matched one.

\subsection{Loss Function}
One training sample is a triplet $(\mathbf{V,Q,A})$, which consists of an input video, an input language query and a ground truth moment set. During training, we need to determine which candidate moments correspond to the ground truth moment and train the network accordingly.

Specifically, for each candidate moment $(t^s_i,t^e_i)$ in the answer set $A$, we first compute the Intersection-over-Union (IoU) score $o_{i}$ between it and ground truth moment $(g^s,g^e)$. Following \cite{zhang2020learning}, we scale the IoU score $o_{i}$ with two thresholds $\theta_{min}$ and $\theta_{max}$, which can be written as:
\begin{equation}
   y_i = 
    \left\{
    \begin{array}{cc}
         0 & o_i \leq \theta_{min} \\
         \frac{o_i - \theta_{min}}{\theta_{max} - \theta_{min}} &  \theta_{min} < o_i < \theta_{max}  \\
         1 & o_i \geq \theta_{max}
    \end{array}
    \right.
\end{equation}
where $y_i$ is used as the supervision label. Finally, we adopt an alignment loss to align the predicted confidence scores with the scaled IoU, which is formulated by:
\begin{equation}
    \mathcal{L} = -\frac{1}{N_A} \Sigma_{i=1}^{N_A}(y_i\log p_i + (1 - y_i)\log(1 - p_i)),
\end{equation}
where $p_i$ is the output score of the answer choice $a_i$ and $N_A$ is the total number of valid candidate moments.

\section{Experiments}

\begin{table*}[t]
    \centering
    \caption{Performance comparison on three benchmarks \ie Charades-STA, TACoS, ActivityNet Captions. \textbf{Note}: The hidden size of our reported model is 256. The first six methods \cite{gao2017tall,liu2018attentive,xu2019multilevel,zhang2019cross,zhang2020learning,wang2020dual} utilize pre-trained VGG features for Charades-STA, while others apply pre-trained I3D features. The top-2 performance values are highlighted by \textbf{bold} and \underline{underline}, respectively.}
    % \footnotesize
    \begin{tabular}{c|c|cccc|cccc|cccc}
    \toprule
    % \hline
    \multirow{3}{*}{\textbf{Methods}} & \multirow{3}{*}{\textbf{Avenue}} & \multicolumn{4}{c|}{\textbf{Charades-STA}} & \multicolumn{4}{c|}{\textbf{TACoS}} & \multicolumn{4}{c}{\textbf{ActivityNet Captions}}\\
    & & \multicolumn{2}{c}{\textbf{Rank@1}}  & \multicolumn{2}{c|}{\textbf{Rank@5}} & \multicolumn{2}{c}{\textbf{Rank@1}}  & \multicolumn{2}{c|}{\textbf{Rank@5}} & \multicolumn{2}{c}{\textbf{Rank@1}}  & \multicolumn{2}{c}{\textbf{Rank@5}}\\
    & &  \textbf{0.5} & \textbf{0.7} & \textbf{0.5} & \textbf{0.7} 
    & \textbf{0.3} & \textbf{0.5} & \textbf{0.3} & \textbf{0.5} 
    & \textbf{0.5} & \textbf{0.7} & \textbf{0.5} & \textbf{0.7} \\
    % \midrule
    \hline
    CTRL  \cite{gao2017tall} & \textit{ICCV '17} & 23.63 & 8.89 & 58.92 & 29.52 & 18.32 & 13.30 & 36.69 & 25.42 & 29.01 & 10.34 & 59.17 & 37.54\\
    
    ACRN  \cite{liu2018attentive} & \textit{SIGIR '18}  &20.26 &7.64 &71.99 &27.79 &19.52 &14.62 &34.97 &24.88 &31.67 &11.25 &60.34 &38.57\\
    
    QSPN  \cite{xu2019multilevel} & \textit{AAAI '19} &35.60 &15.80 &79.40 &45.40 &20.15 &15.23 &36.72 &25.30 &33.26 &13.43 &62.39 &40.78\\
    
    CMIN  \cite{zhang2019cross} & \textit{SIGIR '19}  & - & - & - & - &24.64 &18.05 &38.46 &27.02 &43.40 &23.88 &67.95 &50.73\\
    
    2D-TAN  \cite{zhang2020learning} & \textit{AAAI '20} &39.70 &23.31  &80.32 &51.26 &37.29 &25.32  &57.81 &45.04 &44.05 &27.38  &76.65 &62.26\\

    DPIN  \cite{wang2020dual} & \textit{ACM MM '20} & 47.98 & 26.96  & 85.53 & 55.00 & 46.74 & 32.92  & 62.16 & 50.26 & 47.27 & 28.31  & 77.45 & 60.03\\
    
    \cline{1-6}
    % Ours & 47.98 & 26.96  & 85.53 & 55.00\\
    
    DRN \cite{Zeng_2020_CVPR} & \textit{CVPR '20}  & 53.09 & 31.50 & 89.06 & 60.05 & - & 23.17 & - & 33.36  & 45.45 & 24.39 & 77.97 & 50.30 \\
    
    VSLNet \cite{zhang2020span} & \textit{ACL '20} & 54.19 & 35.22  & - & - & 29.61 & 24.27  & - & - & 43.22 & 26.16  & - & -\\
    
    FIAN \cite{qu2020fine} & \textit{ACM MM '20} & 58.55 & 37.72  & 87.80 & 63.52 & 33.87 & 28.58  & 47.76 & 39.16 & 47.90 & 29.81  & 77.64 & 59.66\\
    
    CPNet \cite{li2021proposal} & \textit{AAAI '21} & 60.27 & 38.74  & - & - &  42.61 & 28.29  & - & - & 40.56 & 21.63  & - & -\\
    
    BPNet \cite{xiao2021boundary} & \textit{AAAI '21} & 50.75 & 31.64  & - & - & 25.96 & 20.96  & - & - & 42.07 & 24.69  & - & -\\
    
    CI-MHA \cite{xiao2021boundary} & \textit{SIGIR '21} & 54.68 & 35.27  & - & - & - & -  & - & - & 43.97 & 25.13  & - & -\\
    
    RaNet \cite{gao2021relation} & \textit{EMNLP '21} & 60.40 & 39.65  & \underline{89.57} & 64.54  & 43.34 & 33.54  & \underline{67.33} & \underline{55.09} & 45.59 & 28.67  & 75.93 & \underline{62.97}\\
    
    SMIN \cite{wang2021structured} & \textit{CVPR '21} & \textbf{64.06} & \underline{40.75}  & 89.49 & \textbf{68.09} & \underline{48.01} & \underline{35.24}  & 65.18 & 53.36  & \textbf{48.46} & \underline{30.34}  & \textbf{81.16} & 62.11\\

    \hline
    Ours & - & \underline{60.82} & \textbf{41.16}  & \textbf{89.77} & \underline{64.73} & \textbf{48.81} & \textbf{36.74}  & \textbf{71.46} & \textbf{59.24} & \underline{47.92} & \textbf{30.47}  & \underline{78.15} & \textbf{63.56}\\ 

    \bottomrule
    \end{tabular}
    \label{tab:performance-3data}
\end{table*}

\subsection{Datasets and Evaluation Metrics}
To verify the effectiveness of our model, we conduct experiments on three popular benchmarks: Charade-STA~\cite{sigurdsson2018charades}, TACoS~\cite{regneri-etal-2013-grounding} and ActivityNet Captions~\cite{krishna2017dense}.\\
\noindent
\textbf{Charade-STA.} It is built on the Charades dataset by \cite{gao2017tall} with annotated language descriptions, which mainly focuses on daily indoor activities. The duration of each video in Charades-STA is 30.59 seconds on average and the average time of video segments lasts 8.22 seconds. There are $12,408$ and $3,720$ query-moment pairs in the training and testing sets respectively.\\
\noindent
\textbf{TACoS.}
 It consists of 127 videos, which are around 5 minutes on average. Videos in TaCoS are collected from cooking scenarios, which describe different activities happened in kitchen room. TACoS is a more challenging dataset due to the long duration of each video and the lack of scene diversity. We follow the same split as \cite{gao2017tall}, which includes $10,146$, $4,589$, $4,083$ query-moment pairs for training, validation, and testing.\\
\noindent
\textbf{ActivityNet Captions.}
It is originally developed for video captioning and contains 20k untrimmed videos with 100k descriptions from YouTube. Videos in ActivityNet Captions are diverse and open, which are around 2 minutes on average. Following public split, we use val\_1 as validation set and val\_2 as testing set, which have $37,417$, $17,505$, and $17,031$ query-moment pairs for training, validation, and testing, respectively.\\
\noindent
\textbf{Evaluation Metric.} We adopt “R@n, IoU=m” as our evaluation metric as \cite{gao2017tall}. It is defined as the percentage of at least one moment in the top “n” selected moments that has IoU with ground truth larger than the threshold “m”.

\subsection{Implementation Details}
\label{sec:details}

\noindent
\textbf{Feature Extractor.} For a fair comparison, we follow previous works \cite{zhang2020learning,Zeng_2020_CVPR} and apply pre-trained C3D \cite{tran2015learning} to encode the videos in TACoS and ActivityNet Captions, while use VGG \cite{simonyan2014very} and I3D \cite{carreira2017quo} features for Charades-STA.  As for word embedding, we utilize pre-trained GloVe \cite{pennington2014glove} to embed each word into 300 dimension vectors as previous solutions \cite{liu2020jointly}.

\noindent
\textbf{Architecture settings.} The number of sampled clips $T$ is set as 64 for Charades-STA and ActivityNet Captions, while 128 for TACoS. In the coarse-grained feature encoder, We adopt 2-layer Bi-GRUs for language encoding as well as video encoding on Charades-STA and ActivityNet Captions, while 3-layer on TACoS. We stack 4 convolution blocks in our choice comparison module, where each block consists of a group convolution layer followed by a batch normalization and \textit{ReLU} function. The group number is set to 32, the kernel size and padding size are 7 and 3, respectively. The size of all hidden states in our model is set to 256 for computation reduction.

\noindent
\textbf{Training settings.} We adopt Adam with learning rate of $1\times 10^{-3}$ for optimization. The batch size is set to 32 for TACoS and 64 for the other two. For all datasets, we trained the model for 15 epochs in total. The thresholds $\theta_{min}$ and $\theta_{max}$ used in our training loss are set to 0.5 and 1.0 for Charades-STA and ActivityNet Captions, while 0.3 and 0.7 for TACoS, which keep the same as \cite{zhang2020learning}. Our model is implemented in PyTorch 1.1.0 with CUDA 10.2.89 and cudnn 7.5.1. All experiments are conducted with 4 GeForce RTX 2080Ti GPUs.

\subsection{Performance Comparison}
We compare our proposed model with some published works on the moment retrieval task:  CTRL \cite{gao2017tall}, ACRN \cite{liu2018attentive}, QSPN \cite{xu2019multilevel}, CMIN \cite{zhang2019cross}, 2D-TAN \cite{zhang2020learning}, DPIN \cite{wang2020dual}, DRN \cite{Zeng_2020_CVPR},  VLSNet \cite{zhang2020span}, FIAN \cite{qu2020fine}, CPNet \cite{li2021proposal}, BPNet \cite{xiao2021boundary}, CI-MHA \cite{yu2021cross}, RaNet \cite{gao2021relation}, SMIN \cite{wang2021structured}.  Among these published works, \cite{Zeng_2020_CVPR, zhang2020span, li2021proposal, yu2021cross} belong to one-stage models which are proposal-free methods, whereas others are proposal-based methods and can be grouped into two-stage models. We report the result of $n \in\{1, 5\}$ with $m \in\{0.3, 0.5\}$ for TACoS, $n \in\{1, 5\}$ with $m \in\{0.5, 0.7\}$ for Charades-STA and ActivityNet Captions, as shown in Table \ref{tab:performance-3data}. 

In general, our method outperforms most recent approaches on three challenging benchmarks and ranks the first or the second across all evaluation metrics. In particular, our model surpasses the state-of-the-arts methods with a large margin on TACoS dataset. Although most videos in TACoS describe long-time cooking activities, whose scenes are slightly varied and sampled moments are visually similar, our model can achieve state-of-the-arts performance even with fewer parameters. For example, we obtain 36.74\% in terms of R@1,IoU=0.5, and have 4.15\% absolute improvements compared with recent method RaNet \cite{gao2021relation} in terms of R@5,IoU=0.5. For Charades-STA, our model achieves the best performance in terms of R@1,IoU=0.7 and R@5,IoU=0.5 with I3D features used in previous works \cite{zhang2020span, xiao2021boundary}. Our MGPN also ranks first among all existing methods except for SMIN \cite{wang2021structured}, which is well-designed but much more sophisticated. For fair comparison, we also evaluate our model with VGG features as previous works did \cite{liu2018attentive, zhang2019cross}, and achieve 27.45\% in terms of R@1.IoU=0.7, which outperforms DPIN \cite{wang2020dual}. As for ActivityNet Captions dataset, our model still demonstrates its competence and improves the performance from 62.97\% to 63.56\% in terms of R@5,IoU=0.7. It is noteworthy that although some results of our MGPN are slightly lower than previous best methods \cite{gao2021relation, wang2021structured}, our MGPN is easier to reproduce and more efficient. It should be noted that the hidden size of our MGPN is set to 256, which is the half of common methods. When we raise our hidden size to 512, we can obtain stronger performance as seen in \ref{sec:hidden-size}. 

The reasons for our solution outperforming the existing methods mainly lie in two folds: (i) Our method takes human reading habits into consideration and integrates reading strategies into our framework. (ii) Our proposed modules can perceive intra-modality and inter-modality information at a multi-granularity level, thus empower our model for better reasoning. 

\subsection{Retrieval Efficiency Comparison}

\begin{table}[]
    \begin{center}
     \caption{Efficiency comparison in terms of parameters (Param.) and video per second (VPS.) on TACoS.
    We only compare our model with two-stage methods for fairness.
    }
    \label{tab:efficiency}
    {
   
    \begin{tabular}{ccc}
    \toprule
      &  \textbf{Param.} & \textbf{VPS.}  \\
    \hline
    CTRL  & 22M & 4.3    \\
    ACRN & 128M & 2.1  \\
    2D-TAN & 60.93M & 18.33  \\
    RaNet & 12.80M & 21.79  \\
    Ours & \textbf{6.62M} & \textbf{26.88}  \\
    \bottomrule
    \end{tabular}}
    \end{center}
\end{table}
Existing two-stage methods usually suffer from high computation complexity due to densely sampled moments, while our MGPN alleviates this drawback to some extent with the help of smaller hidden size and efficient comparison module. To demonstrate its retrieval efficiency, we compare our MGPN with other published works on TACoS dataset. All experiments are conducted with 4 GeForce RTX 2080Ti GPUs. We report Param. and VPS. of each model in Table \ref{tab:efficiency}. 
“Param.” denotes the total parameters of each model, and “VPS.” denotes the number of videos each model can process per second. 
It can be observed that our MGPN can process most videos within the same time while has least parameters, which demonstrates that our MGPN is an efficient and lightweight retrieval model. 

\subsection{Ablation Study}

\subsubsection{Effectiveness of each component}
\label{sec:component}

\begin{table}[t]
	\centering{
	\caption{ Effectiveness of each component in our proposed MGPN on ActivityNet Captions, where "sec-enc.", "sec-fuse." and "comp." denote fine-grained feature encoders in \ref{sec:method-fine-encoder}, conditioned interaction module in \ref{sec:method-fine-fuse} and choice comparison module in \ref{sec:method-comparison}, respectively. \cmark \ or \xmark \ means the net with or without the component}
	\label{tab:ablation-components}
	
	\begin{tabular}{cccccc}
    \toprule
     \multirow{2}{*}{\textbf{Model}}& \multicolumn{3}{c}{\textbf{Components}} & \multicolumn{2}{c}{\textbf{Rank@1}}\\
      & \textbf{sec-enc.} & \textbf{sec-fuse.} & \textbf{comp.} & \textbf{0.5} & \textbf{0.7}\\
    \midrule
   \textcircled{1}  & \xmark&\xmark&\xmark& 44.33 & 24.83\\
   \textcircled{2}  & \cmark&\xmark&\xmark&  45.21 & 27.03 \\
   \textcircled{3}  & \cmark&\cmark&\xmark& 46.21 & 27.87\\
   \textcircled{4}  & \xmark&\xmark&\cmark& 46.64 & 28.75\\
   \textcircled{5}  & \cmark&\cmark&\cmark&\textbf{47.92} & \textbf{30.47}\\
    
    \bottomrule
	\end{tabular}%
	}

\end{table}%

% \textcolor{red}{\textbf{TODO:}}
To evaluate the effectiveness of each component in our MGPN, we conduct in-depth ablation studies as shown in Table \ref{tab:ablation-components}. Model \textcircled{1} is our baseline model which directly applies a choice ranker over the generated moment features. It can achieve acceptable performance due to preliminary understanding obtained by our coarse-grained feature encoder and co-attention module. Model \textcircled{2} adds a fine-grained feature encoder to capture detailed information in intra-modality and boosts the retrieval performance. It indicates that we need to encode video content as well as sentence context at a multi-granularity level for more precise retrieval, just like we need to read passage and question once again for correct selection in reading comprehension. Model \textcircled{3} further adds a conditioned interaction module based on Model \textcircled{2} to fully mine inter-modality information. The performance improvement demonstrates the effectiveness of fine-grained alignment between videos and sentences. Model \textcircled{4} adds a choice comparison module over baseline model \textcircled{1} and obtains satisfying results, demonstrating the importance of temporal relations among candidate moments. Model \textcircled{5} is our full MGPN whose performance surpasses all the ablation models.
% integrates fine-grained feature encoder, conditioned interaction module and choice comparison module. 
From Table \ref{tab:ablation-components} we can observe that all the proposed components in our MGPN can bring obvious performance improvements. We can conclude that the reading strategies (\ie passage question reread, enhanced passage question alignment, and choice comparison) integrated into our framework are meaningful and effective.

\subsubsection{Investigation on the feature encoder}
\label{sec:experiment-encoder}
\begin{table}[h]
	\centering{
	\caption{Effectiveness of different feature encoders on ActivityNet Captions.  w/. means "with". }
	\label{tab:feature-encoder}
	\begin{tabular}{c|cccc}
    \toprule
     \multirow{2}{*}{\textbf{Encoders}} & \multirow{2}{*}{\textbf{VPS.}} & \multirow{2}{*}{\textbf{Param.}} & \multicolumn{2}{c}{\textbf{Rank@1}}\\
      &  &  & \textbf{0.3} & \textbf{0.5} \\
    \hline
   w/.GRU  & \textbf{89.94} & \textbf{5.12M}  & 47.92 & 30.47 \\
   w/.LSTM  & 84.17 & 5.53M  & 47.27 & 29.84 \\
   w./Transformer  & 24.42 & 16.91M  &  \textbf{48.69} & \textbf{30.59}  \\
    
    \bottomrule
	\end{tabular}%
	}
	
\end{table}%
 Transformer block \cite{vaswani2017attention} is a proven strong feature encoder which can exhaustively capture the long range dependencies of sequence features \cite{zhang2020span, wang2021self, zhang2021video}. 
 To investigate the effect of different feature encoders, we replace the Bi-GRU in our coarse-grained encoder with a Transformer block. Considering that previous methods \cite{zhang2020learning, xiao2021boundary, gao2021relation} tend to adopt LSTM \cite{hochreiter1997long} as the text encoder, we also report the result with the Bi-LSTM encoder. As shown in Table \ref{tab:feature-encoder}, Bi-GRU is more suitable for contextual perception in our solution compared with Bi-LSTM. Transformer feature extractor is beneficial for slight performance improvements while it leads to lower speed and larger parameters. For the sake of both retrieval performance and computation cost, we choose the lightweight Bi-GRU as our feature encoder. 

\begin{figure}[t]
    \centering
    \includegraphics[width=1.0\linewidth]{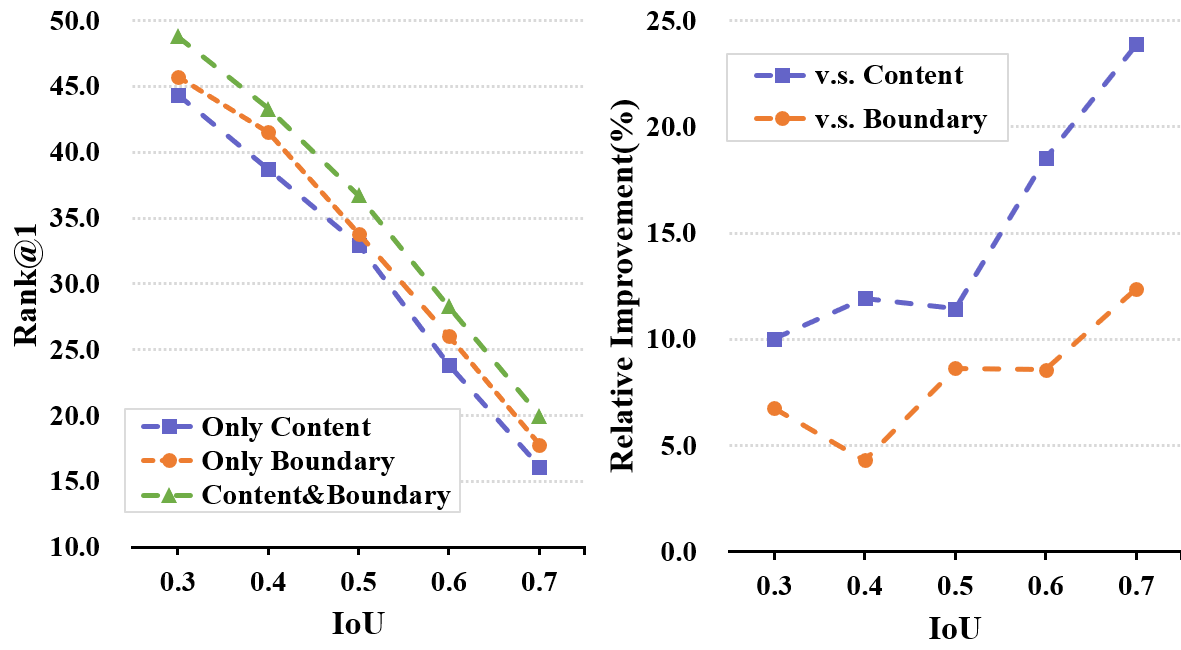}
    \caption{Left: Performance comparison of different ways to generate candidate moment feature maps. Right: Relative improvements achieved by combining both content-level and boundary-level representations. Experiments are conducted on TACoS. (best viewed in color) }
    \label{fig:fig-generator-compare}
\end{figure}

\subsubsection{Investigation on the choice representation generator}
\label{sec:generator-compare}
In our choice representation generation module, we take both content-level and boundary-level representations into account for comprehensive perception. We use \textit{MaxPooling} operation to generate content-level moment features $\mathbf{A_C}$ and \textit{Addition} operation for boundary-level moment features $\mathbf{A_B}$ generation. Content-level feature $\mathbf{A_C}$ is responsible to capture context within the temporal span of the candidate moment, while boundary-level feature $\mathbf{A_B}$ aims to making cross-modal retrieval boundary sensitive. We also investigate the effect of different choice representation generators as shown in Figure \ref{fig:fig-generator-compare}. We can see from the illustration that combination of both $\mathbf{A_C}$ and $\mathbf{A_B}$ outperforms only using $\mathbf{A_C}$ or $\mathbf{A_B}$, and larger relative improvements are obtained for higher IoUs. The reasonable results indicate that both content-level and boundary-level representations play an important role in moment retrieval. It can be observed that considering only boundary-level features surpasses only content-level features, which is in line with the case in RaNet \cite{gao2021relation}. We speculate that moment retrieval is a boundary sensitive task and thus the impact of boundary information is crucial.

\begin{figure}[t]
    \centering
    \includegraphics[width=1.0\linewidth]{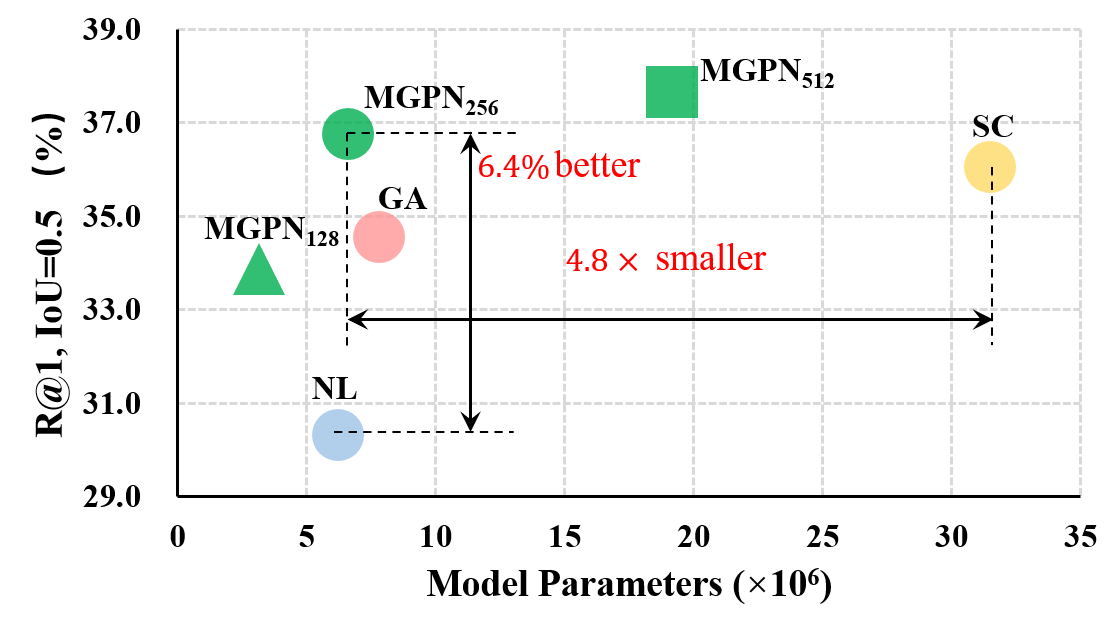}
    \caption{Performance comparison on TACoS in terms of R@1,IoU=0.5 and model parameters. SC, GA, and NL denote stacked convolution, graph attention and non-local, respectively. MGPN$_{128}$, MGPN$_{256}$ and MGPN$_{512}$ represent MGPN with hidden size 128, 256, 512, respectively. 
    % (best viewed in color) 
    }
    \label{fig:fig-comparison}
\end{figure}

\subsubsection{Investigation on the comparison module}
\label{sec:experiment-choice-comparison}

For model efficiency and easy implementation consideration, we apply a series of stacked convolution blocks to learn moment-wise temporal relations, where each block consists of a group convolution layer followed by a batch normalization and \textit{ReLU} activation function. To demonstrate the effectiveness of our choice comparison module, we replace our stacked convolution blocks with other implementations for better comparison. Particularly, we conduct experiments on stacked convolution layers (SC) used in 2D-TAN \cite{zhang2020learning}, graph attention layer (GA) proposed in RaNet \cite{gao2021relation}, and non-local blocks (NL) \cite{wang2018non}. We send the enhanced fused features $\hat{\mathbf{A}}$ discussed in section \ref{sec:method-comparison} to above modules respectively to obtain relation-aware feature $\Tilde{\mathbf{A}}$. Results are illustrated in Figure \ref{fig:fig-comparison}, it can be observed that our stacked convolution blocks (\ie MGPN$_{256}$) can achieve excellent performance with relatively fewer parameters. Stacked convolution blocks proposed in our MGPN$_{256}$ is 6.4\% better than non-local blocks (NL) implementation and 4.8$\times$ smaller than stacked convolution layers (SC) used in 2D-TAN \cite{zhang2020learning}.

\subsubsection{Investigation on the hidden sizes}
\label{sec:hidden-size}

Existing two-stage methods tend to suffer from large model size, thus it is a matter of concern to design an lightweight architecture to alleviate this drawback. Most previous works \cite{gao2017tall, zhang2020learning, qu2020fine} set 512 as the hidden size of their architecture without further discussion. We conduct experiments on different hidden sizes to investigate whether large hidden size is necessary. In our paper, we choose 256 as our hidden size and denote the model as MGPN$_{256}$. We further introduce two variant models MGPN$_{128}$ and MGPN$_{512}$, whose hidden size are 128 and 512, respectively. The comparative study on their performance is shown in \ref{fig:fig-comparison}. It is observed that although models with larger hidden size can achieve slight performance improvement, their model size increases exponentially. Therefore, we finally set the hidden size as 256 in our implementation to reach a balance between performance and efficiency.

\subsubsection{Plug-and-play}
\label{sec:plug}

The core of our proposed MGPN is the fine-grained perception of intra-modality as well as inter-modality information. We do experiments on several published works to demonstrate the feasibility and generalization of our motivation. Concretely, we select two open-source two-stage models, \ie 2D-TAN \cite{zhang2020learning} and RaNet \cite{gao2021relation}. Both of them encode video and sentence features in a coarse manner and integrate them only once, which neglect the reread human habits. To embody our motivation on these two models, we maintain their own feature encoders, moment-query interaction module and temporal adjacent network (or called as moment comparison module). Then we append a video-query alignment module and a fine-grained feature encoder before their own interaction module. The added alignment module is our proposed multi-modal co-attention module discussed in section \ref{sec:method-coarse-fuse}, and the added feature encoder is our fine-grained feature encoder proposed in section \ref{sec:method-fine-encoder}. Results are reported in Table \ref{tab:plug}, where "\textit{Once}" means we coarsely read video passage and query question to directly select an answer, while "\textit{Twice}" means we read again to fully understand and integrate video and query content. From the reported results we can conclude that appending fine-grained self-modal encoding and cross-modal alignment are beneficial for better retrieval.

\begin{table}[]
	\centering{
	\caption{Plug-and-play experiments on ActivityNet Captions. "Once" denotes original results reported in their papers. "Twice" denotes the models considering fine-grained perception.}
	\label{tab:plug}
	\begin{tabular}{c|cc|cc}
    \toprule
     \multirow{2}{*}{\textbf{Methods}} & \multicolumn{2}{c|}{\textbf{Once}} & \multicolumn{2}{c}{\textbf{Twice}}\\
      &  \textbf{R@1,0.5} & \textbf{R@1,0.7} &  \textbf{R@1,0.5} & \textbf{R@1,0.7} \\
    \hline
   2D-TAN  & 37.29 & 25.32 & \textbf{43.06} & \textbf{26.92} \\

   RaNet &  45.59 & 28.67 & \textbf{46.46} & \textbf{29.78} \\
   MGPN &  46.64 & 28.75 & \textbf{47.92} & \textbf{30.47}\\
    
    \bottomrule
	\end{tabular}%
	}
	
\end{table}%

\subsection{Qualitative Analysis}
\label{sec:visualize}
\begin{figure}
    \centering
    \subfigure[Examples on the ActivityNet Captions dataset]{
    \includegraphics[width=1.0\linewidth]{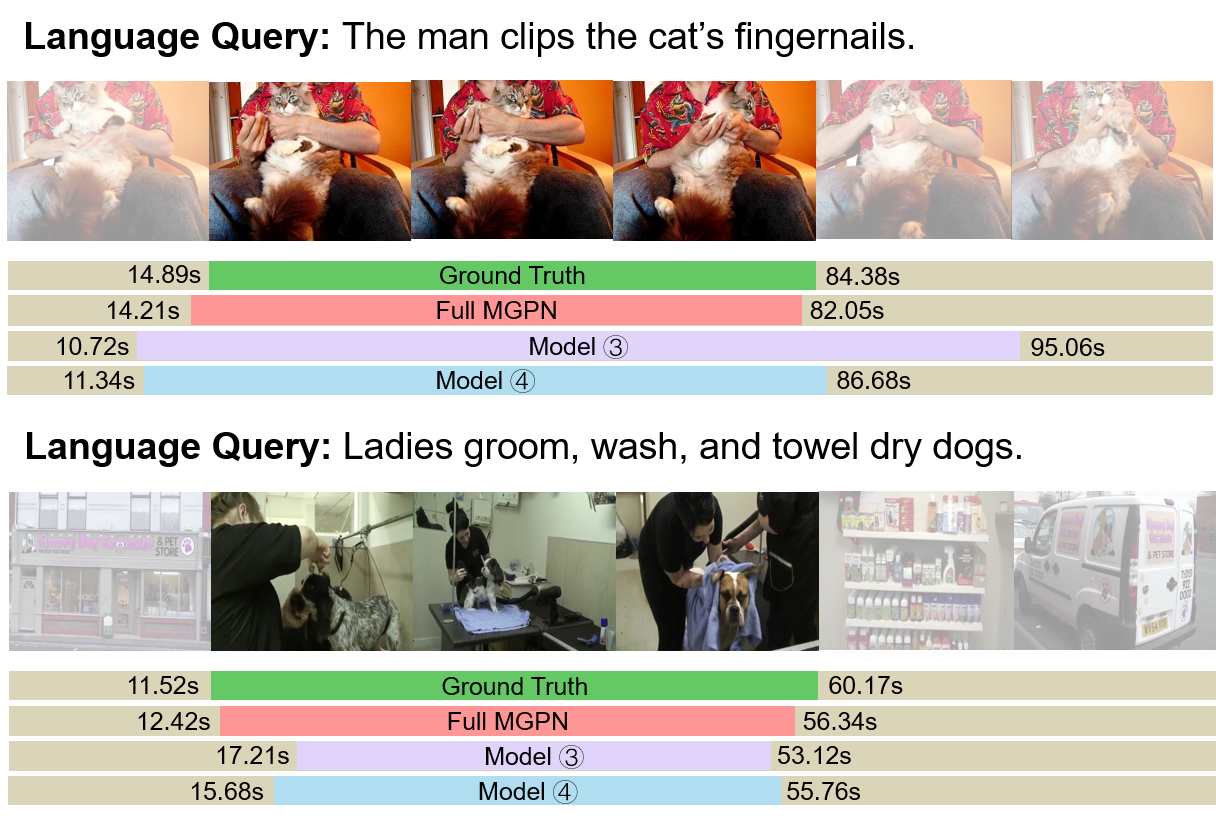}
    \label{fig:quality-analysis1}
    }
    \subfigure[Examples on the TACoS dataset]{
    \includegraphics[width=1.0\linewidth]{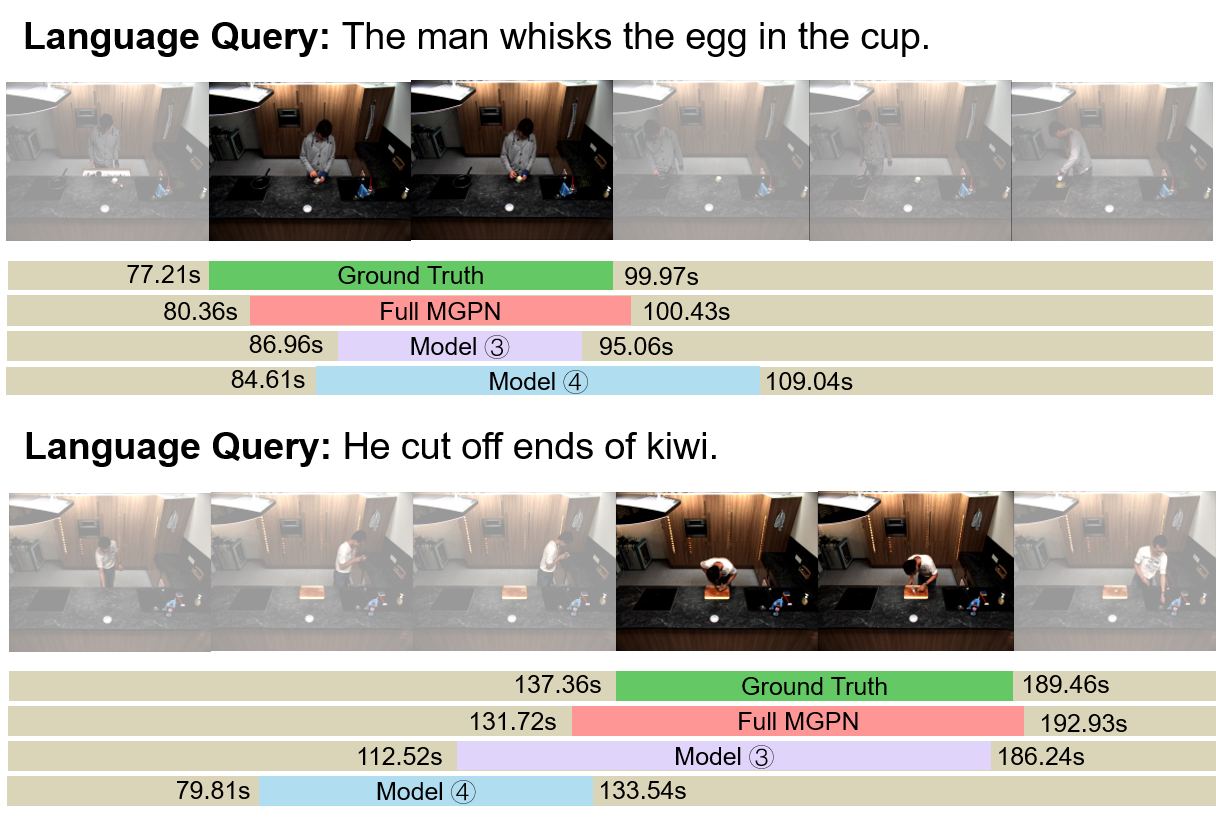}
    \label{fig:quality-analysis2}
    }
    \caption{Qualitative examples of our MGPN and ablation models evaluated on the ActivityNet Captions and TACoS dataset (best viewed in color).}
    \label{fig:quality-analysis}
\end{figure}

To demonstrate the effectiveness of MGPN, we further provide qualitative analysis on ActivityNet Captions and TACoS dataset. As shown in Figure \ref{fig:quality-analysis}, MGPN is capable of precisely retrieving the moment most relevant to the language query, even though those moments are visually similar. We also illustrate the qualitative results of Model \textcircled{3} and Model \textcircled{4} discussed in \ref{sec:component}. In general, the results reflect that fine-grained intra-modality and inter-modality learning are crucial for moment retrieval, because they are in line with human reading habits thus can prompt the model to make more precise retrieval.

\section{Conclusion}
In this paper, we formulate moment retrieval task from the perspective of multi-choice reading comprehension and propose a novel \textbf{M}ulti-\textbf{G}ranularity \textbf{P}erception \textbf{N}etwork (\textbf{MGPN}) to tackle it. We integrate several human reading strategies (\ie passage question reread, enhanced passage question alignment, choice comparison) into our framework and accordingly design the fine-grained feature encoder, conditioned interaction and moment comparison module. These modules empower our model to perceive intra-modality and inter-modality information at a multi-granularity level for better reasoning. Extensive experiments on Charades-STA, TACoS and ActivityNet Captions datasets have demonstrated the effectiveness and efficiency of our proposed MGPN. We will devote efforts to a more effective fine-grained video encoder in future work.

% \input{sections/6-acknowledgement}

% \bibliographystyle{ACM-Reference-Format}
% \bibliography{reference.bbl}

\end{document}